\documentclass[pdflatex,sn-mathphys-num]{sn-jnl}

\usepackage{threeparttable}%
\usepackage{graphicx}%
\usepackage{multirow}%
\usepackage{amsmath,amssymb,amsfonts}%
\usepackage{amsthm}%
\usepackage{mathrsfs}%
\usepackage[title]{appendix}%
\usepackage{xcolor}%
\usepackage{colortbl}
\definecolor{rowpink}{HTML}{FFDBEA}
\definecolor{rowgray}{HTML}{EAEAEA}
\usepackage{textcomp}%
\usepackage{manyfoot}%
\usepackage{booktabs}%
\usepackage{tabularx}
\usepackage{algorithm}%
\usepackage{algorithmicx}%
\usepackage{algpseudocode}%
\usepackage{listings}%
\usepackage{makecell}
\usepackage{array}
\usepackage{hyperref} 
\usepackage{url} 
\usepackage{mathrsfs}
\newcolumntype{Y}{>{\centering\arraybackslash}X}
\usepackage{caption} 

\theoremstyle{thmstyleone}%
\theoremstyle{thmstyletwo}%

\theoremstyle{thmstylethree}%

\begin{document}

\title{A report-grounded vision--language foundation model for colonoscopy from 280{,}000 routine reports}



\author[1,2,4,8]{\fnm{Jia} \sur{Yu}}
\equalcont{These authors contributed equally to this work.}

\author[2,5]{\fnm{Yan} \sur{Zhu}}
\equalcont{These authors contributed equally to this work.}

\author[1,2]{\fnm{Yili} \sur{He}}

\author[8]{\fnm{Zilong} \sur{Wang}}

\author[8]{\fnm{Xinyang} \sur{Jiang}}

\author[2,5]{\fnm{Peiyao} \sur{Fu}}

\author[4]{\fnm{Ruijie} \sur{Yang}}

\author[2]{\fnm{Tianyi} \sur{Chen}}

\author[2]{\fnm{Siyuan} \sur{Li}}

\author[3,4]{\fnm{Zhihua} \sur{Wang}}

\author[3,4]{\fnm{Fei} \sur{Wu}}

\author*[2,5]{\fnm{Quanlin} \sur{Li}}\email{li.quanlin@zs-hospital.sh.cn}

\author*[6,7]{\fnm{Xian} \sur{Yang}}\email{xian.yang@manchester.ac.uk}

\author*[2,5]{\fnm{Pinghong} \sur{Zhou}}\email{zhou.pinghong@zs-hospital.sh.cn}

\author*[1,2,7]{\fnm{Shuo} \sur{Wang}}\email{shuowang@fudan.edu.cn}


\affil[1]{%
  \orgname{Digital Medical Research Center, School of Basic Medical Sciences, Fudan University},
  \orgaddress{\city{Shanghai}, \country{China}}}

\affil[2]{%
  \orgname{Shanghai Collaborative Innovation Center of Endoscopy},
  \orgaddress{\city{Shanghai}, \country{China}}}
  
\affil[3]{%
  \orgname{Zhejiang University},
  \orgaddress{\city{Hangzhou}, \country{China}}}

\affil[4]{%
  \orgname{Shanghai Institute for Advanced Study of Zhejiang University},
  \orgaddress{\city{Shanghai}, \country{China}}}

\affil[5]{%
  \orgdiv{Endoscopy Centre and Endoscopy Research Institute},
  \orgname{Zhongshan Hospital, Fudan University},
  \orgaddress{\city{Shanghai}, \country{China}}}

\affil[6]{%
  \orgname{Alliance Manchester Business School, The University of Manchester},
  \orgaddress{\city{Manchester}, \country{UK}}}

\affil[7]{%
  \orgname{Data Science Institute, Imperial College London},
  \orgaddress{\city{London}, \country{UK}}}
  
\affil[8]{%
  \orgname{Microsoft Research Asia},
  \orgaddress{\city{Shanghai}, \country{China}}}

\abstract{Vision--language models remain underused in colonoscopy despite the rich expert descriptions recorded in routine reports. These reports document lesion appearance, size and location but summarise entire procedures rather than caption individual frames, leaving clinical findings only weakly linked to the corresponding images. Here we develop EndoCLIP, a colonoscopy vision--language foundation model trained on 125{,}756 lesion-level image--text pairs progressively recovered from 280{,}476 routine colonoscopy records. Across lesion-level image--text retrieval, structured report generation and six multi-centre clinical classification tasks, EndoCLIP outperforms general-purpose and biomedical vision--language encoders in both zero-shot and linear-probe settings. On benign-versus-malignant classification, its linear probe approaches the performance of expert readers in a blinded study involving 12 endoscopists. These results suggest that recovering finding-to-frame correspondence can transform routine documentation into scalable supervision, enabling clinical targets to be specified in language rather than separately annotated for each task.}




\maketitle

\section{Introduction}\label{intro}

\begin{figure}[b]
\centering
\includegraphics[width=\textwidth]{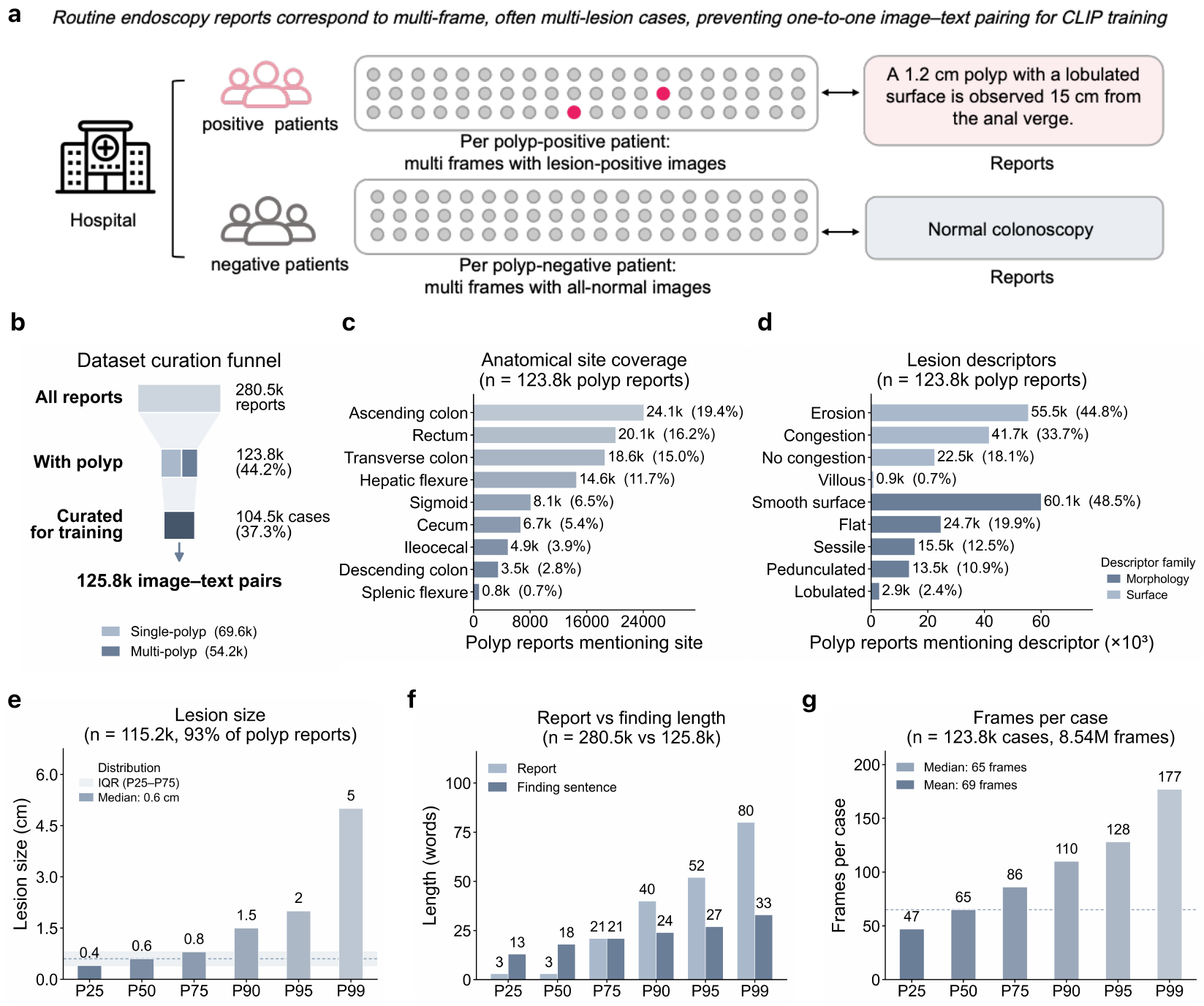}
\caption{\textbf{Weak report--image alignment and analysis of the pretraining dataset.} \textbf{a}, Routine reports correspond to multi-frame, often multi-lesion cases. \textbf{b}, Dataset curation funnel, from 280.5k de-identified reports to 123.8k polyp-positive reports and a curated training set of 104.5k cases (37.3\%) yielding 125.8k image--text pairs. \textbf{c}, Anatomical-site coverage across the 123.8k polyp reports. \textbf{d}, Frequency of morphology and surface descriptors. \textbf{e}, Lesion-size distribution. \textbf{f}, Report and finding-sentence length in words. \textbf{g}, Frames per case (n~=~123.8k cases, 8.54M frames; median 65, mean 69). In \textbf{e}--\textbf{g}, bars mark distribution percentiles (P25--P99).}
\label{fig:dataset}
\end{figure}

Colonoscopy is central to the prevention of colorectal cancer, a leading cause of cancer-related mortality worldwide \cite{bray2024global,winawer1993prevention,zauber2012colonoscopic,bretthauer2022nordicc}. AI-assisted systems have improved adenoma detection and supported the automated detection and classification of colorectal lesions \cite{urban2018deep,ahmad2019artificial,wang2019real,repici2020efficacy}. However, most such systems rely on task-specific models trained with densely annotated and carefully curated images, which is difficult to scale across the diverse findings encountered in routine colonoscopy. Their robustness and generalisability across patient populations, imaging systems and clinical centres also remain concerns \cite{parasa2023framework,boers2024foundation,11080481}.

Self-supervised visual pretraining can reduce this dependence on manual annotation \cite{zhou2023retfound}. Visual foundation models such as GastroNet \cite{boers2024foundation} and EndoFM \cite{wang2023endofm} learn transferable visual features from large collections of unlabelled endoscopy frames and videos. These features are not linked to clinical language and therefore do not directly support text-guided retrieval or prompt-based recognition. Vision--language pretraining instead aligns images with text, allowing clinical concepts to be specified in natural language \cite{radford2021learning,jia2021scaling,cherti2023reproducible}. In radiology and pathology, this approach has been successful in zero-shot and label-efficient recognition \cite{zhang2022convirt,huang2021gloria,tiu2022chexzero,huang2023plip,lu2024conch,hamamci2026ctclip}, while in colonoscopy it remains less explored \cite{wang2025endoked}. Routine colonoscopy reports describe lesion morphology, surface appearance, size and anatomical location but summarise the examination rather than caption individual frames. A procedure may contain dozens of still frames, many without a clear lesion view, and one report may describe several lesions without linking findings to frames. The report and visual record are therefore only coarsely aligned, especially in multi-lesion procedures (Fig.~\ref{fig:dataset}).

\begin{figure}[htbp]
\centering
\includegraphics[width=1.0\textwidth]{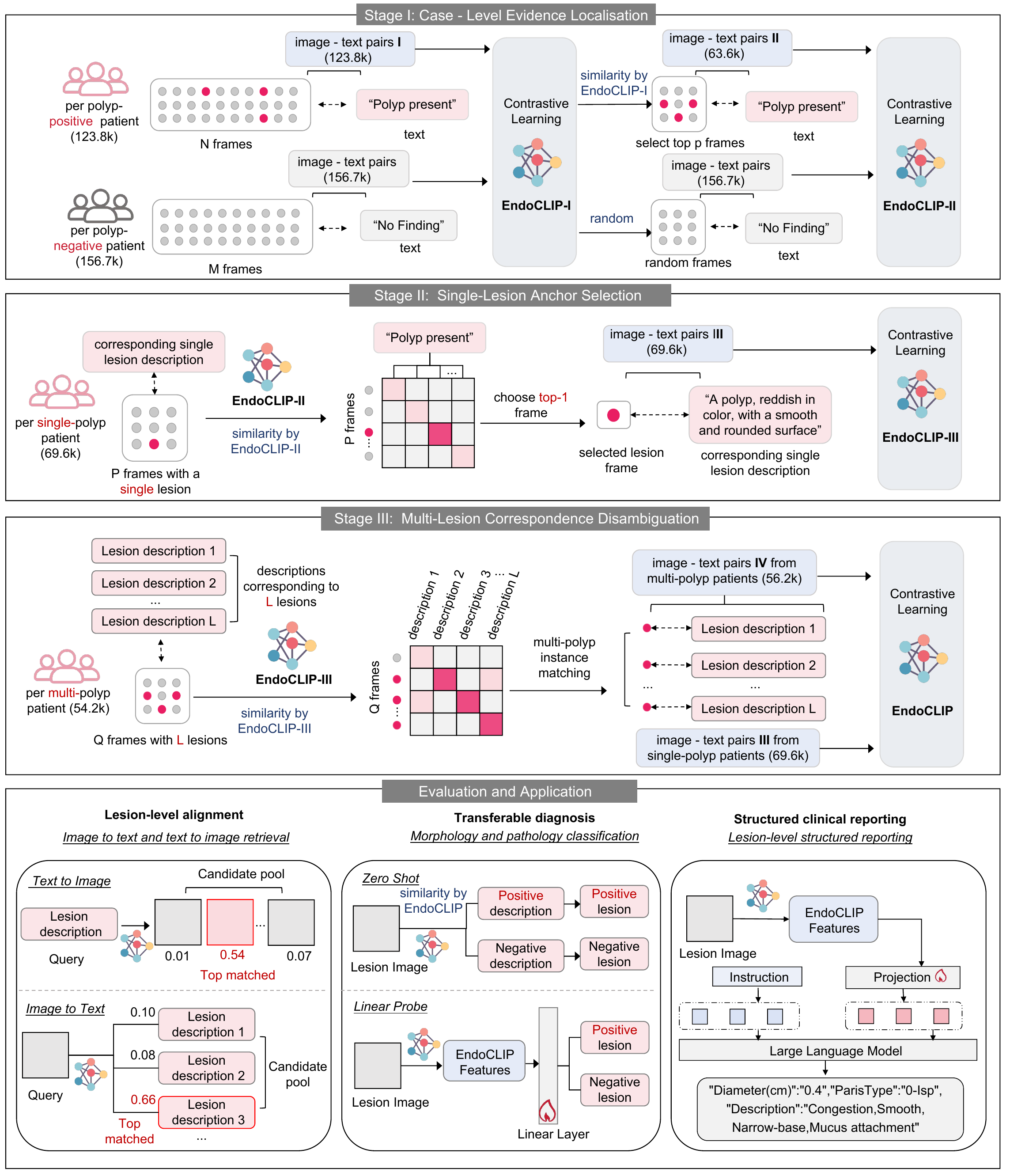}
\caption{\textbf{Overview of EndoCLIP.} Routine reports describe multi-frame, sometimes multi-lesion procedures rather than individual images. EndoCLIP derives lesion-level image--text pairs from this case-level signal through case-level evidence localisation, single-lesion anchor selection and multi-lesion disambiguation. The resulting encoder is evaluated on retrieval, classification and structured report generation.}
\label{fig:framework}
\end{figure}

Here we introduce EndoCLIP, a vision--language foundation model that recovers finding-to-frame correspondence through case-level evidence localisation, single-lesion anchor selection and multi-lesion disambiguation. Applied to 280{,}476 routine reports, this approach yields 125{,}756 lesion-level image--text pairs for contrastive pretraining (Fig.~\ref{fig:dataset}). We also introduce EndoReport100 \cite{li2026endoreport100}, a clinician-curated benchmark of 100 multi-lesion cases and 7{,}002 frame-level annotations for evaluating lesion-level retrieval. EndoCLIP achieves a bidirectional global Recall@1 of 14.3\%, compared with 2.8\% for the strongest comparator. Furthermore, we validated the performance of EndoCLIP on six clinical classification tasks and structured report generation in a multi-centre setting~\cite{fu2026endovl}. Its frozen features achieve the highest zero-shot and label-efficient classification performance among the compared vision--language encoders and yield the most accurate structured outputs. These results suggest that recovering finding-to-frame correspondence can turn routine reports into scalable supervision for clinically relevant endoscopy tasks with little or no task-specific annotation.

\section{Results}
\label{sec:results}

\subsection{Overview of EndoCLIP}
\label{sec:overview}
\label{sec:weak_alignment}
\label{sec:framework}

EndoCLIP learns from routine free-text reports generated during colonoscopy. Because each report documents an entire examination rather than individual frames, its finding sentences are not explicitly linked to the frames that show the corresponding lesions (Fig.~\ref{fig:dataset}). To recover this correspondence, we used a three-stage progressive training procedure that localises case-level visual evidence, selects anchors from single-lesion cases and disambiguates finding-to-frame matches in multi-lesion reports (Fig.~\ref{fig:framework}; Methods, Sec.~\ref{subsec:correspondence_recovery}). Applied to 280{,}476 reports and their still frames, this procedure yields 125{,}756 lesion-level image--text pairs for contrastive pretraining (Supplementary Table~S1). Contrastive training with coarse case-level prompts (``polyp present'' or ``no finding'') produces EndoCLIP-I, which localises polyp-containing frames. Retraining on the retained high-scoring frames yields EndoCLIP-II, which filters uninformative background. Matching finding sentences to frames in single-lesion cases with EndoCLIP-II produces EndoCLIP-III, the first checkpoint to encode descriptive lesion semantics. Disambiguating multi-lesion cases with EndoCLIP-III and combining all recovered pairs then produces EndoCLIP, the completed model evaluated below; the intermediate checkpoints are used only in stage-wise analyses.

We benchmark EndoCLIP against general-purpose and biomedical vision--language encoders across lesion-level retrieval (Fig.~\ref{fig:retrieval}), six clinical classification tasks spanning a multi-centre public benchmark and an independent pathology cohort (Fig.~\ref{fig:cls} and Table~\ref{tab:size}), embedding-space organisation (Fig.~\ref{fig:tsne}) and structured report generation (Fig.~\ref{fig:report}).

\subsection{Report-grounded lesion retrieval}
\label{sec:retrieval}

\begin{figure}[htbp]
\centering
\includegraphics[width=\textwidth]{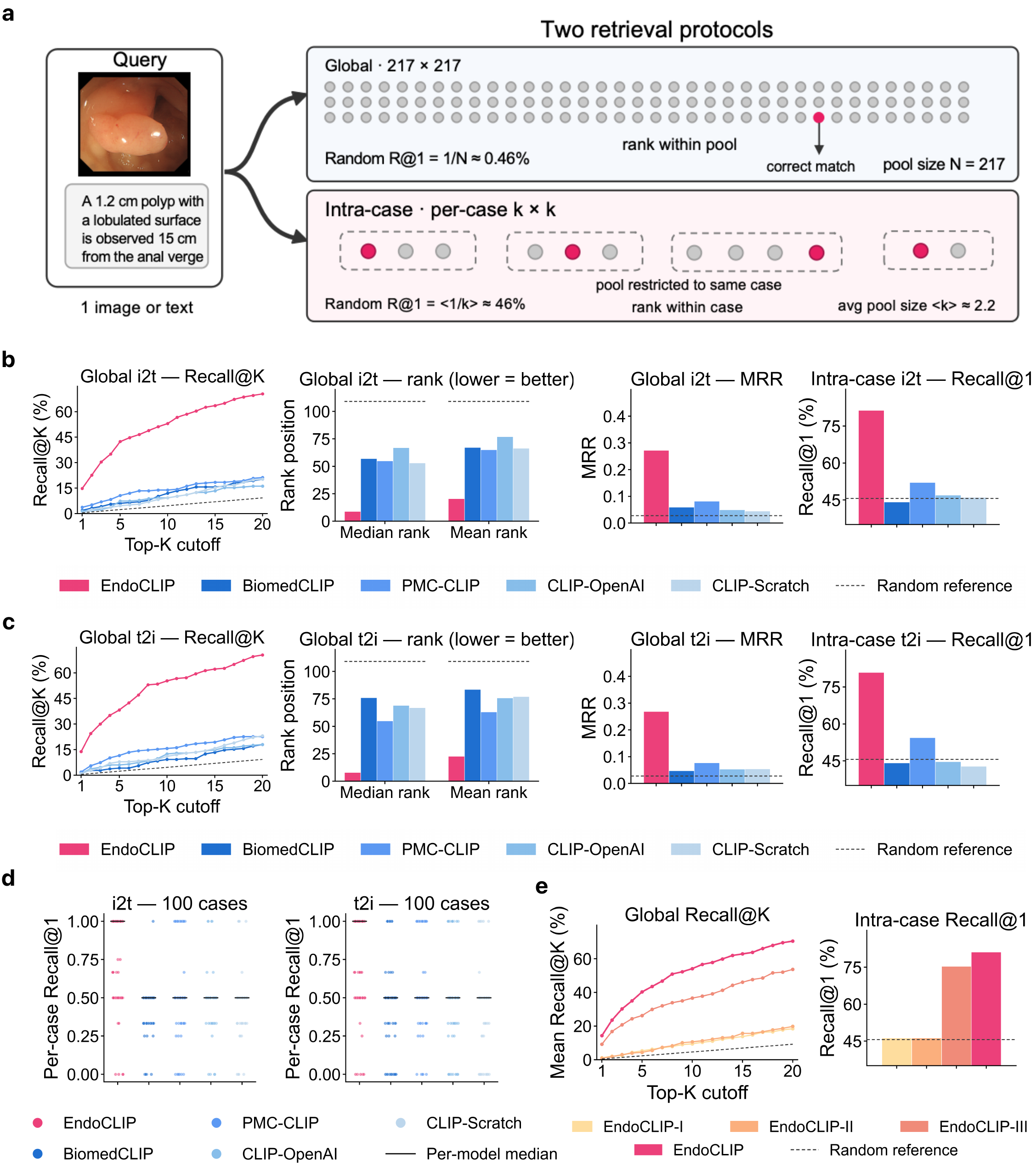}
\caption{\textbf{Lesion-level image--text retrieval.} \textbf{a}, Global and intra-case protocols. \textbf{b},\textbf{c}, Image-to-text (\textbf{b}) and text-to-image (\textbf{c}) retrieval, each showing Recall@K curves, rank statistics, mean reciprocal rank (MRR) and intra-case Recall@1. \textbf{d}, Per-case intra-case Recall@1 across all multi-lesion cases. \textbf{e}, Performance across successive EndoCLIP training stages.}
\label{fig:retrieval}
\end{figure}

Lesion-level retrieval tests whether EndoCLIP encodes lesion-specific clinical semantics in its shared representation. We use EndoReport100, a clinician-curated benchmark of 100 multi-lesion cases and 217 lesions that is fully independent of the pretraining corpus. Experienced endoscopists marked polyp presence in each frame and linked every lesion to a representative frame and its finding sentence (Fig.~\ref{fig:retrieval}a). The global protocol compares each query with all candidates, whereas the intra-case protocol restricts candidates to lesions documented in the same case (mean, 2.2 per case).

General-purpose and biomedical encoders show limited lesion-level retrieval. Under the global protocol, where chance Recall@1 is 0.46\%, PMC-CLIP~\cite{lin2023pmcclip} reaches a mean Recall@1 of 2.8\% (95\% CI 1.2--4.7), and the general-purpose CLIP variants perform similarly to the randomly initialised network. EndoCLIP reaches a mean Recall@1 of 14.3\% (95\% CI 10.1--18.4) across the two retrieval directions and reduces the median matched-item rank to 8--9, compared with 53--76 for the baselines. The paired difference from PMC-CLIP is 11.5 percentage points ($p<0.001$, case-level bootstrap). The same ordering is observed under the intra-case protocol, which has a 46\% chance level. PMC-CLIP reaches 53.3\% (95\% CI 48.7--58.1), and EndoCLIP reaches 81.2\% (95\% CI 75.9--86.0; $p<0.001$; Fig.~\ref{fig:retrieval}b--d; full results in Supplementary Table~S3 and Supplementary Fig.~S1).

Successive checkpoints reveal when descriptive retrieval emerges. EndoCLIP-I and EndoCLIP-II, trained only on the Stage~I ``polyp present'' and ``no finding'' prompts, detect polyp-containing frames but do not support descriptive lesion retrieval. They reach frame-level AUROCs of 0.843 and 0.889 on the 7{,}002 labelled EndoReport100 frames (Supplementary Fig.~S4 and Supplementary Tables~S16--S17). Their global mean Recall@1 values on descriptive retrieval are 1.2\% and 0.9\%, respectively, and their intra-case Recall@1 is 46.3\% (Fig.~\ref{fig:retrieval}e). EndoCLIP-III, the first checkpoint trained on selected single-lesion sentence--frame pairs, reaches a global mean Recall@1 of 9.2\% (95\% CI 5.7--13.1) and an intra-case Recall@1 of 75.3\% (95\% CI 70.0--80.6). Relative to EndoCLIP-III, the completed EndoCLIP increases global and intra-case Recall@1 by 5.1 percentage points ($p=0.02$) and 5.8 percentage points ($p=0.003$), respectively. Full values are given in Supplementary Table~S4 and Supplementary Fig.~S2; the case-representative protocol shows the same ordering (Supplementary Table~S19). A control encoder trained on the same corpus with naive case-level pairing, in which each case contributes one randomly matched frame--sentence pair, reaches a global mean Recall@1 of 6.5\% (95\% CI 3.9--9.5) and an intra-case Recall@1 of 70.2\% (95\% CI 63.5--76.7), below the completed model by 7.8 and 10.9 percentage points, respectively (both $p<0.001$; Supplementary Table~S21).

\subsection{Multi-centre zero-shot and label-efficient recognition}

\begin{figure}[htbp]
\centering
\includegraphics[width=0.99\textwidth]{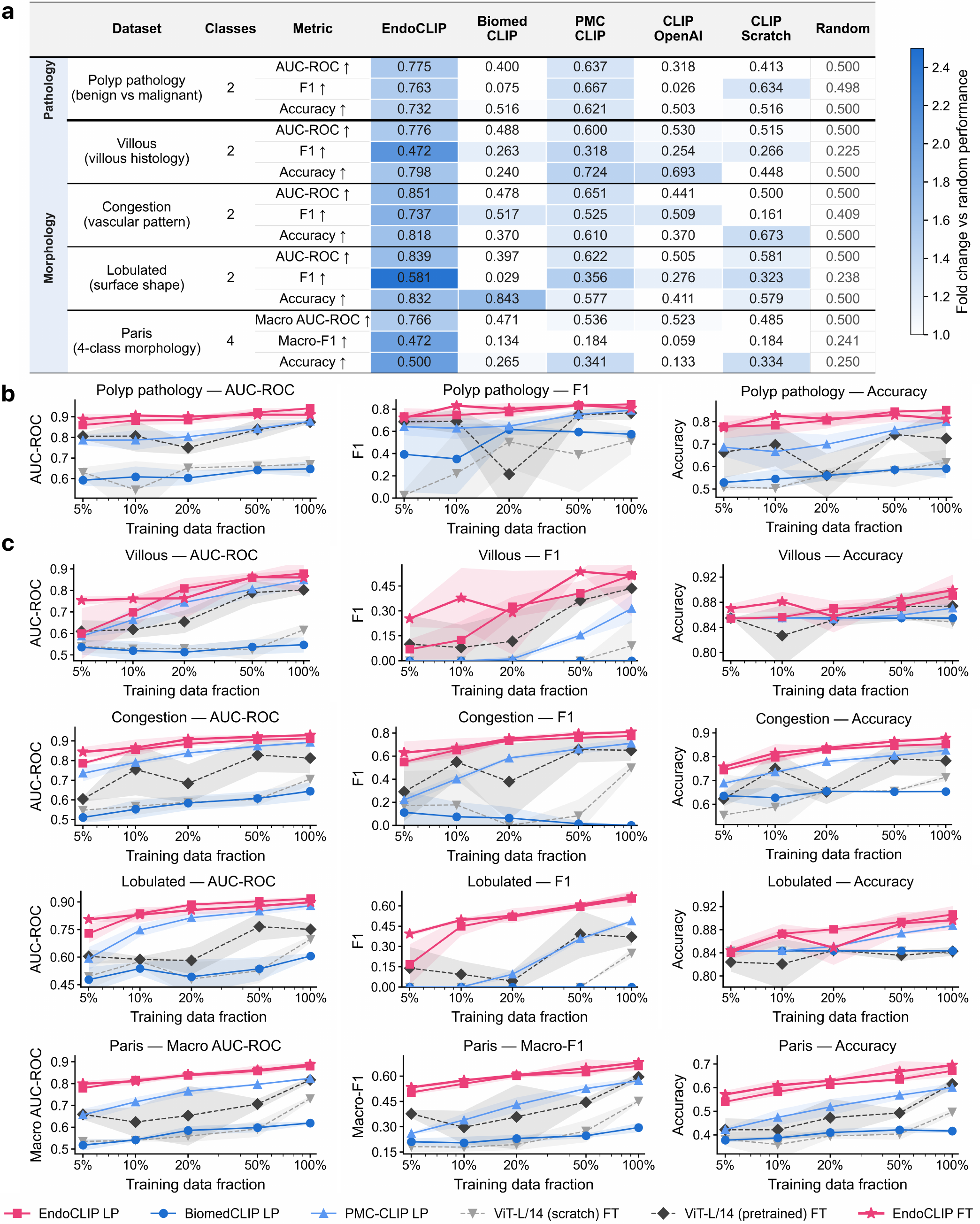}
\caption{\textbf{Multi-centre zero-shot and label-efficient clinical classification.} Five tasks are shown: four EndoVL morphology and surface tasks \cite{fu2026endovl} and pathology from Zhongshan Hospital; polyp size is reported in Table~\ref{tab:size}. \textbf{a}, Prompt-based zero-shot performance, with colour encoding fold change over the random baseline. \textbf{b},\textbf{c}, AUC-ROC, F1 and accuracy across five training-data fractions under linear probing and end-to-end training. Lines and shaded bands show the mean and s.d. over three seeds.}
\label{fig:cls}
\end{figure}

We evaluate EndoCLIP across six clinical tasks that combine a multi-centre external benchmark with an independent pathology cohort. Five tasks come from EndoVL \cite{fu2026endovl}, which spans nine public source datasets: four-class Paris morphology, mucosal congestion, lobulated surface, villous surface and polyp size. The sixth is benign-versus-malignant classification on a Zhongshan Hospital cohort. These targets carry direct clinical weight: Paris type and surface pattern inform invasion risk and resection planning \cite{paris2003endoscopic,hayashi2013nice}, lesion size at the 10~mm threshold sets the surveillance interval \cite{gupta2020recommendations}, and the benign--malignant distinction separates endoscopic from surgical management. All six tasks are assessed under three regimes: prompt-based zero-shot scoring, linear probing on frozen encoders and end-to-end training, with the two supervised regimes using 5--100\% of the training partition (Fig.~\ref{fig:cls}; Table~\ref{tab:size}; Supplementary Tables~S2 and S13--S14).

Without task-specific fitting, EndoCLIP achieves the highest AUC-ROC among the compared encoders on every classification task. Across the external morphology and surface tasks, AUC-ROC is 0.766--0.851, exceeding PMC-CLIP~\cite{lin2023pmcclip} by 0.176--0.230; on in-house pathology, it is 0.775, a margin of 0.138 (Fig.~\ref{fig:cls}a; Supplementary Fig.~S3 and Supplementary Table~S7). For external polyp-size classification at the 10~mm threshold, EndoCLIP reaches 0.670, compared with 0.567 for the strongest comparator (Table~\ref{tab:size}a).

On lobulated surface, BiomedCLIP~\cite{zhang2025biomedclip} attains a marginally higher accuracy than EndoCLIP (0.843 versus 0.832) with a lower sensitivity and F1 (0.015 and 0.029 versus 0.746 and 0.581). Class-imbalanced predictions also occur for BiomedCLIP on mucosal congestion (sensitivity 0.972, specificity 0.052) and for PMC-CLIP on polyp size (sensitivity 0.002, specificity 0.998). Across the external binary tasks, EndoCLIP has a sensitivity between 0.623 and 0.746 and a specificity between 0.827 and 0.863 (Supplementary Table~S7).

\begin{table}[b]
\centering
\captionsetup{width=\textwidth}
\caption{\textbf{Label-efficient polyp-size classification on the multi-centre EndoVL benchmark.} Polyps are dichotomised at the clinically established 10~mm (1~cm) threshold that separates small from large lesions~\cite{gupta2020recommendations}. The biomedical comparators are BiomedCLIP~\cite{zhang2025biomedclip} and PMC-CLIP~\cite{lin2023pmcclip}. \textbf{(a)}~Zero-shot prediction from clinical prompts. \textbf{(b)}~Label-efficient transfer at five training-data fractions under two regimes: a linear probe on the frozen image encoder and end-to-end training with all weights updated; accuracy and AUC-ROC are reported as the mean$\pm$s.d.\ over three seeds. End-to-end training comprises full fine-tuning of EndoCLIP and pretrained ViT-L/14 and training ViT-L/14 from scratch. Within each regime the best entry per column is shown in \textbf{bold} and the second best is \underline{underlined}.}
\label{tab:size}
\begin{minipage}{\textwidth}
\footnotesize
\centering
\setlength{\tabcolsep}{2pt}

\begin{tabularx}{\linewidth}{@{}lYYYYYY@{}}
\multicolumn{7}{@{}l}{\textbf{(a) Zero-shot} \textit{(text prompts, no training labels)}}\\[2pt]
\toprule
\textbf{Method} & \textbf{Acc.} & \textbf{AUROC} & \textbf{F1} & \textbf{Sens.} & \textbf{Spec.} & \textbf{Prec.} \\
\midrule
BiomedCLIP & 0.556 & \underline{0.567} & \underline{0.533} & \underline{0.585} & 0.534 & 0.491 \\
PMC-CLIP & \underline{0.565} & 0.378 & 0.004 & 0.002 & \textbf{0.998} & \underline{0.500} \\
CLIP-OpenAI & 0.556 & 0.531 & 0.377 & 0.310 & \underline{0.745} & 0.483 \\
\rowcolor{rowpink} EndoCLIP & \textbf{0.624} & \textbf{0.670} & \textbf{0.631} & \textbf{0.739} & 0.536 & \textbf{0.550} \\
\bottomrule
\end{tabularx}

\vspace{1.5ex}

\begin{tabularx}{\linewidth}{@{}llYYYYY@{}}
\multicolumn{7}{@{}l}{\textbf{(b) Label-efficient transfer} \textit{(accuracy / AUC-ROC, mean$\pm$s.d.)}}\\[2pt]
\toprule
\textbf{Method} & \textbf{Metric} & \textbf{5\%} & \textbf{10\%} & \textbf{20\%} & \textbf{50\%} & \textbf{100\%} \\
\midrule
\multicolumn{7}{@{}l}{\textit{Linear probe (frozen image encoder)}} \\
 & Acc.\ & 0.692$\pm$0.021 & \underline{0.726$\pm$0.029} & \underline{0.760$\pm$0.016} & \underline{0.758$\pm$0.017} & \underline{0.769$\pm$0.017} \\
\multirow{-2}{*}{BiomedCLIP} & AUC & \underline{0.783$\pm$0.009} & \underline{0.792$\pm$0.016} & \underline{0.827$\pm$0.006} & \underline{0.833$\pm$0.003} & 0.844$\pm$0.006 \\
\addlinespace[2pt]
 & Acc.\ & \underline{0.694$\pm$0.015} & 0.697$\pm$0.045 & 0.754$\pm$0.010 & 0.743$\pm$0.018 & 0.763$\pm$0.007 \\
\multirow{-2}{*}{PMC-CLIP} & AUC & 0.764$\pm$0.031 & 0.761$\pm$0.043 & 0.820$\pm$0.008 & 0.831$\pm$0.006 & \underline{0.845$\pm$0.008} \\
\addlinespace[2pt]
 & Acc.\ & 0.608$\pm$0.050 & 0.673$\pm$0.037 & 0.716$\pm$0.002 & 0.732$\pm$0.015 & 0.757$\pm$0.021 \\
\multirow{-2}{*}{CLIP-OpenAI} & AUC & 0.716$\pm$0.008 & 0.735$\pm$0.042 & 0.801$\pm$0.013 & 0.818$\pm$0.006 & 0.841$\pm$0.014 \\
\addlinespace[2pt]
\rowcolor{rowpink}  & Acc.\ & \textbf{0.832$\pm$0.013} & \textbf{0.847$\pm$0.022} & \textbf{0.869$\pm$0.025} & \textbf{0.878$\pm$0.014} & \textbf{0.882$\pm$0.026} \\
\rowcolor{rowpink} \multirow{-2}{*}{EndoCLIP} & AUC & \textbf{0.908$\pm$0.020} & \textbf{0.922$\pm$0.028} & \textbf{0.934$\pm$0.023} & \textbf{0.941$\pm$0.020} & \textbf{0.946$\pm$0.019} \\
\midrule
\multicolumn{7}{@{}l}{\textit{End-to-end training (all weights updated)}} \\
 & Acc.\ & 0.606$\pm$0.030 & 0.625$\pm$0.023 & 0.656$\pm$0.035 & 0.718$\pm$0.015 & 0.742$\pm$0.008 \\
\multirow{-2}{*}{ViT-L/14 (scratch)} & AUC & 0.627$\pm$0.021 & 0.638$\pm$0.028 & 0.717$\pm$0.038 & 0.777$\pm$0.009 & 0.813$\pm$0.014 \\
\addlinespace[2pt]
\rowcolor{rowgray}  & Acc.\ & \underline{0.761$\pm$0.013} & \underline{0.816$\pm$0.011} & \underline{0.875$\pm$0.019} & \underline{0.903$\pm$0.011} & \textbf{0.914$\pm$0.022} \\
\rowcolor{rowgray} \multirow{-2}{*}{ViT-L/14 (pretrained)} & AUC & \underline{0.822$\pm$0.004} & \underline{0.885$\pm$0.009} & \underline{0.929$\pm$0.013} & \underline{0.953$\pm$0.010} & \textbf{0.960$\pm$0.012} \\
\addlinespace[2pt]
\rowcolor{rowpink}  & Acc.\ & \textbf{0.851$\pm$0.014} & \textbf{0.865$\pm$0.018} & \textbf{0.904$\pm$0.013} & \textbf{0.908$\pm$0.013} & \underline{0.913$\pm$0.019} \\
\rowcolor{rowpink} \multirow{-2}{*}{EndoCLIP} & AUC & \textbf{0.924$\pm$0.016} & \textbf{0.933$\pm$0.018} & \textbf{0.958$\pm$0.009} & \textbf{0.955$\pm$0.010} & \underline{0.959$\pm$0.009} \\
\bottomrule
\end{tabularx}
\end{minipage}
\end{table}

\begin{figure}[htbp]
\centering
\includegraphics[width=0.95\textwidth]{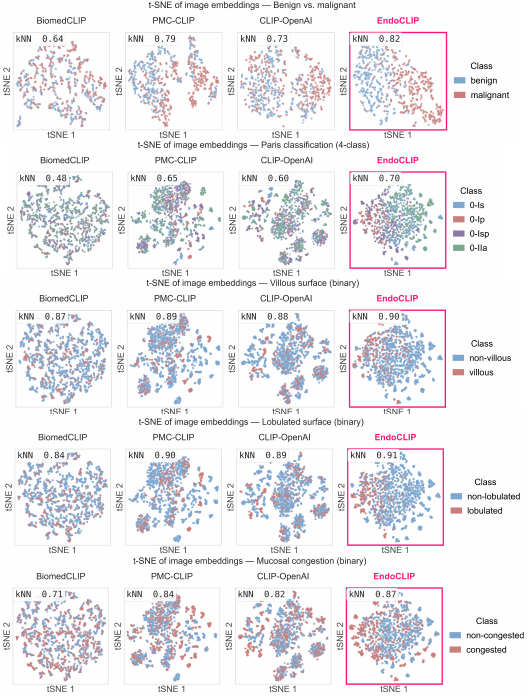}
\caption{\textbf{Clinical organisation of the embedding space.} Rows show t-SNE projections of frozen image embeddings for pathology, Paris morphology, villous surface, lobulated surface and congestion; columns compare four encoders. Points are coloured by ground-truth class, and panels report kNN leave-one-out accuracy. The morphology and surface tasks use EndoVL; pathology is in-house.}
\label{fig:tsne}
\end{figure}

Linear probes on frozen EndoCLIP features have the highest mean AUC-ROC at every label budget. With 5\% of labels, AUC-ROC reaches 0.860 on pathology and 0.908 on polyp size; with the full training partitions, it reaches 0.878--0.941 across morphology, surface and pathology and 0.946 on polyp size (Fig.~\ref{fig:cls}b,c; Table~\ref{tab:size}b; Supplementary Tables~S8--S10 and S14). The margin is widest where labels are scarcest. On polyp size, the EndoCLIP probe fitted to 5\% of labels already exceeds every comparator probe fitted to the full labelled set (0.908 versus at most 0.845), and on pathology and Paris morphology the 5\%-label probe likewise exceeds the CLIP-OpenAI probe fitted to all labels (0.860 versus 0.767 and 0.780 versus 0.725). The ordering is already present in the frozen features themselves: EndoCLIP has the highest k-nearest-neighbour leave-one-out accuracy on every task, from 0.702 to 0.906 against 0.646 to 0.898 for PMC-CLIP, the two models being closest on lobulated and villous surface, where the gap is 0.008 (Fig.~\ref{fig:tsne}; Supplementary Table~S12).

Frozen comparator features also struggle to separate the minority classes. On villous and lobulated surface, probes on BiomedCLIP and CLIP-OpenAI features stay at the majority-class solution at every label fraction, with an F1 of at most 0.011, and PMC-CLIP reaches only 0.314 and 0.488 with the full labelled set. EndoCLIP probes are the only ones to produce a non-trivial minority-class F1 at the smallest budgets, reaching 0.069 and 0.168 with 5\% of labels and 0.515 and 0.655 with all labels (Supplementary Tables~S9 and S10). The largest margins occur on villous and lobulated surface, the two rarest descriptors in the pretraining corpus, present in 0.7\% and 2.4\% of polyp reports (Fig.~\ref{fig:dataset}d). Both are closely associated with advanced histology~\cite{gupta2020recommendations}.

Under end-to-end training, fully fine-tuned EndoCLIP leads the morphology, surface and pathology comparisons in mean AUC-ROC at every label fraction. Its AUC-ROC ranges from 0.754 to 0.888 at 5\% of labels and from 0.860 to 0.930 at 100\% (Fig.~\ref{fig:cls}b,c). On polyp size, fully fine-tuned EndoCLIP performs best through the 50\% label fraction and is similar to the pretrained ViT-L/14 baseline with all labels (0.959 versus 0.960).

To relate model performance to clinical readers, we compare EndoCLIP with 12 endoscopists from three experience tiers who were blinded to the histopathological reference (Supplementary Table~S15). Reader accuracy ranges from 0.716$\pm$0.014 for novices to 0.846$\pm$0.014 for experts. The frozen EndoCLIP probe reaches an accuracy of 0.852$\pm$0.010, an AUC-ROC of 0.941$\pm$0.002 and a specificity of 0.896, compared with a mean reader specificity of 0.704. Exploratory McNemar tests give $p=0.86$ against the expert vote and $p<0.05$ against the novice-tier vote and each external encoder.

\label{sec:embedding_structure}

\begin{figure}[b]
\centering
\includegraphics[width=\textwidth]{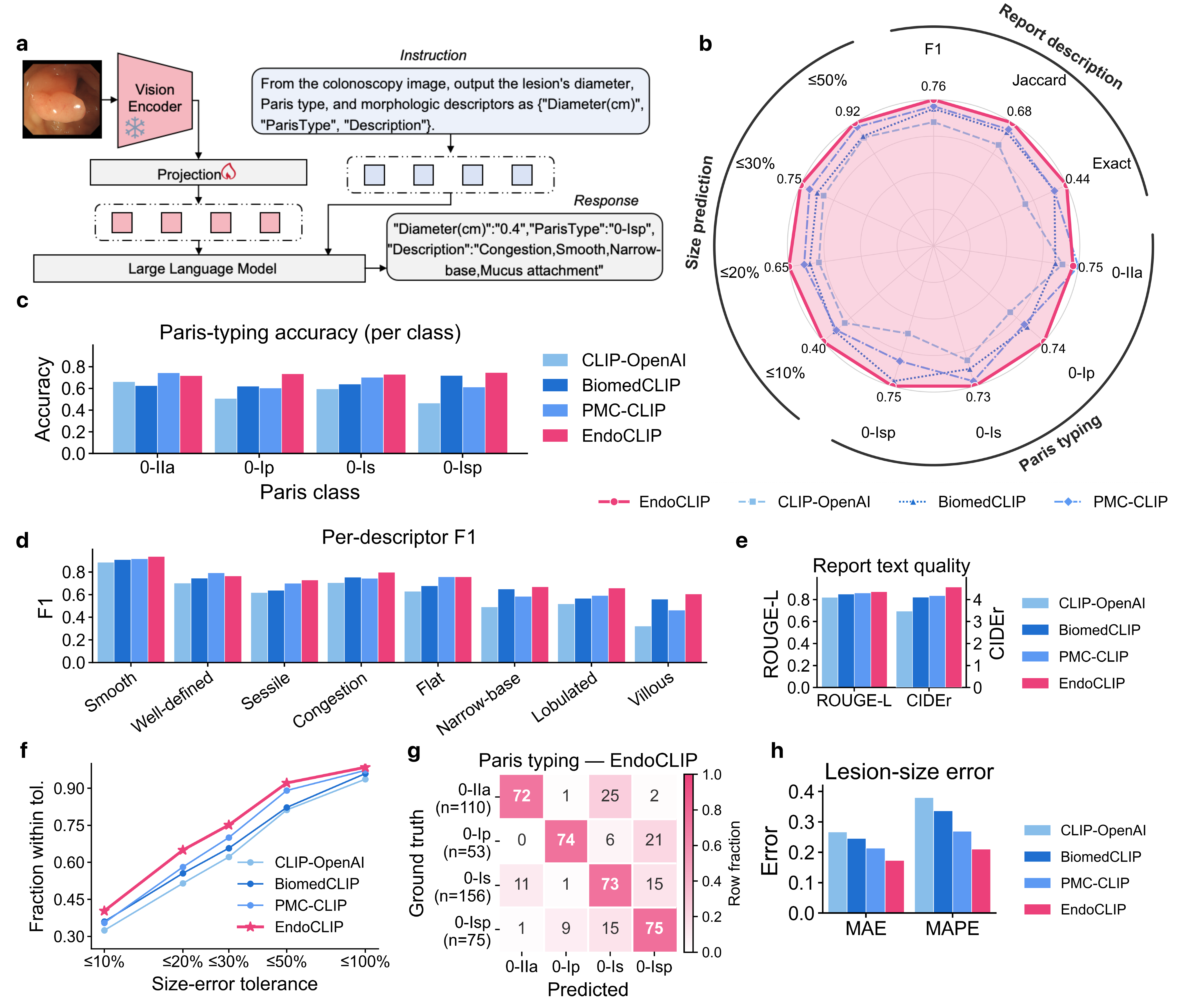}
\caption{\textbf{Structured report generation from frozen features.} \textbf{a}, A frozen vision encoder is connected by a trainable multilayer projector to a frozen Qwen3-14B decoder. \textbf{b}, Summary radar over descriptor-set agreement, Paris typing and size tolerance. Each spoke is scaled to the best-performing encoder and labelled with that best native value; only EndoCLIP is shaded. \textbf{c}, Per-class Paris typing accuracy. \textbf{d}, Per-descriptor F1 for eight findings. \textbf{e}, Report text-quality scores (ROUGE-L and CIDEr). \textbf{f}, Fraction of size estimates within increasing relative-error tolerances. \textbf{g}, Row-normalised EndoCLIP confusion matrix for Paris typing; rows are ground truth and the diagonal reproduces the per-class accuracies in \textbf{c}. \textbf{h}, Lesion-size mean absolute error (MAE, cm) and mean absolute percentage error (MAPE). The vision encoder and decoder are frozen; a separate projector with the same architecture and training protocol is fitted for each encoder. Held-out EndoVL test set, $n=394$.}
\label{fig:report}
\end{figure}

\subsection{Structured report generation from frozen visual features}
\label{sec:clinical_generation}

Structured report generation tests whether frozen representations retain clinically reportable information. Each image encoder is connected through a trainable multilayer projector to the same frozen Qwen3-14B decoder (Fig.~\ref{fig:report}a). Given a fixed instruction, the decoder produces JSON containing lesion diameter, Paris type and descriptors from a 19-term vocabulary. All comparisons use the same projector design and training protocol, with separate projector weights fitted to accommodate different encoder dimensions. The EndoVL structured-report set comprises 1{,}375 training, 197 validation and 394 held-out test images. We compare CLIP-OpenAI, BiomedCLIP~\cite{zhang2025biomedclip}, PMC-CLIP~\cite{lin2023pmcclip} and EndoCLIP on the test set.

EndoCLIP produces the most accurate structured outputs among the encoders compared here (Fig.~\ref{fig:report}b--d; Supplementary Tables~S5 and S6). Its descriptor-set F1 is 0.764, with a Jaccard index of 0.683 and an exact-set-match rate of 0.442; the corresponding values are 0.729, 0.647 and 0.401 for PMC-CLIP and 0.716, 0.631 and 0.401 for BiomedCLIP. EndoCLIP has the highest F1 on six of the eight most frequent descriptors, and its largest gains fall on the two findings that the comparators recover least well: villous surface (0.607, against 0.562 for BiomedCLIP and 0.463 for PMC-CLIP) and lobulated surface (0.660, against 0.593 for PMC-CLIP). PMC-CLIP remains ahead on well-defined margins (0.794 versus 0.766) and is level on flat morphology (0.759 versus 0.758).

Paris typing follows the same pattern. Overall accuracy is 0.731 for EndoCLIP, 0.685 for PMC-CLIP, 0.650 for BiomedCLIP and 0.579 for CLIP-OpenAI, and EndoCLIP is the most accurate encoder on three of the four classes, with per-class accuracy between 0.718 and 0.747 and PMC-CLIP ahead only on 0-IIa (0.746 versus 0.718; Fig.~\ref{fig:report}c). Among the misclassified images, 91\% are assigned to an adjacent category on the protrusion continuum (Fig.~\ref{fig:report}g). Flat-elevated lesions are typed as sessile in 25\% of cases and sessile lesions as flat-elevated in 11\%, pedunculated lesions are typed as sub-pedunculated in 21\%, and sub-pedunculated lesions divide between their sessile and pedunculated neighbours (15\% and 9\%). 

EndoCLIP also gives the closest lesion-size estimates. Its mean absolute error is 0.173~cm, against 0.214~cm for PMC-CLIP, 0.246~cm for BiomedCLIP and 0.267~cm for CLIP-OpenAI, and its mean relative error is 0.211, against 0.269, 0.337 and 0.380 (Fig.~\ref{fig:report}h). Predicted size falls within 10\%, 20\% and 50\% of the reference for 40.4\%, 65.0\% and 92.1\% of lesions, compared with 35.5\%, 58.1\% and 89.1\% for PMC-CLIP, and EndoCLIP leads at every tolerance (Fig.~\ref{fig:report}f). Text-similarity scores give the same ranking, with CIDEr, ROUGE-L and Levenshtein similarity of 4.573, 0.873 and 0.956 against 4.182, 0.861 and 0.951 for PMC-CLIP (Fig.~\ref{fig:report}e). 

\section{Discussion}\label{sec:discussion}

Vision--language pretraining remains uncommon in gastrointestinal endoscopy. Most endoscopy foundation models learn from images or videos and are evaluated on predefined visual tasks \cite{wang2023endofm,boers2024foundation}. In radiology and pathology, by contrast, paired clinical text supports retrieval, zero-shot recognition and language generation \cite{lu2024conch,hamamci2026ctclip}. EndoCLIP extends this paradigm to colonoscopy by learning jointly from routine reports and procedure frames. The resulting representation supports retrieval from clinical descriptions, prompt-defined recognition, adaptation to local labels and structured reporting. Clinical language can therefore provide a common interface across tasks that otherwise need separate annotations and output heads.

Routine reports, however, do not provide the image--caption pairs used in standard CLIP training \cite{radford2021learning,zhang2025biomedclip,lin2023pmcclip}. The unit of documentation is the procedure, whereas the desired unit of cross-modal learning is a lesion-level frame--sentence pair. A report can summarise many frames and several lesions, so direct case-level pairing risks associating a finding sentence with another lesion or with normal mucosa. EndoCLIP narrows this many-to-many ambiguity progressively, beginning with case-level prompts that localise relevant evidence. Single-lesion reports then provide low-ambiguity descriptive anchors for sentence-to-frame disambiguation in multi-lesion cases. This procedure converts case-level weak supervision into finding-to-frame correspondences for cross-modal training.

The checkpoint sequence separates coarse polyp detection from report-grounded lesion retrieval. EndoCLIP-I and EndoCLIP-II reach frame-level AUROCs of 0.843 and 0.889. Despite this, their global descriptive Recall@1 remains 1.2\% and 0.9\%, and their intra-case Recall@1 remains at the 46.3\% chance level. Descriptive training on single-lesion anchors raises these values to 9.2\% globally and 75.3\% within cases for EndoCLIP-III. Adding disambiguated multi-lesion pairs further raises them to 14.3\% and 81.2\%, gains of 5.1 and 5.8 percentage points, respectively. Stage-wise classification shows the largest later-stage gains on morphology-rich tasks when 5\% of labels are available (Supplementary Table~S11). These results separate coarse frame-level lesion detection from descriptive lesion retrieval, which emerges after training on lesion-level sentence--frame pairs.

The shared image--text space also changes how a downstream task is specified. An image-only encoder generally requires labelled examples and a task-specific classifier for each new target. EndoCLIP can instead score clinical prompts directly, while retaining the option of a linear probe or end-to-end adaptation when local labels are available. It achieves the highest zero-shot AUC-ROC among the compared encoders on all six classification tasks and exceeds the strongest comparator by 0.176--0.230 across the four external morphology and surface tasks. These targets include Paris morphology, surface appearance and lesion size, which inform invasion risk, resection strategy and surveillance \cite{paris2003endoscopic,hayashi2013nice,gupta2020recommendations}. This capability may be relevant to rare or newly defined phenotypes for which large labelled cohorts are difficult to assemble \cite{lu2024conch,hamamci2026ctclip}.

Structured report generation poses a stricter test because a clinically useful report must reflect the image rather than merely sound plausible. Previous multi-centre studies have developed image-based structured reporting and prospectively evaluated domain-specific multimodal models for upper-gastrointestinal diagnosis and reporting \cite{qu2020reportgeneration,jiang2026reportangel}. GI-Bench shows that readable outputs from general-purpose multimodal language models may still contain localisation errors, factual inaccuracies and hallucinated visual findings \cite{zhu2026gibench}. With the decoder frozen, EndoCLIP yields the most accurate descriptor sets, Paris types and lesion-size estimates among the encoders compared. These results highlight the value of endoscopy-specific image--text data and visually grounded encoders beyond language-model fluency alone, and support structured drafts for endoscopist review as a clinically relevant use of endoscopy-specific vision--language models.

The potential clinical value of these capabilities lies in assisting endoscopists across several parts of the workflow. Language-based retrieval could support case review and teaching by locating lesions from textual descriptions, extending the image-query paradigm used in EndoFinder \cite{Yan_EndoFinder_MICCAI2024}. Prompt-based recognition and lightweight local adaptation could support quality assurance when annotation is limited, while structured outputs could reduce repetitive documentation. EndoReport100 provides a benchmark for developing frame-level detection and report-grounded retrieval in multi-lesion procedures. Evaluation on EndoVL and the independent Zhongshan Hospital pathology cohort provides evidence of transfer across heterogeneous data sources beyond the pretraining corpus \cite{fu2026endovl}. On pathology, the frozen probe reaches an accuracy close to the expert-tier point estimate in the reader study. Prospective workflow studies can now determine whether these capabilities improve clinical efficiency or decision-making.

Several features of the current evidence constrain the scope of these conclusions. Lesion sizes in the pretraining corpus are parsed from report text rather than measured directly, and EndoReport100, although independent of pretraining, comes from the same institutional archive. Patient- or video-level separation cannot be verified for the EndoVL report-generation partitions because those identifiers are unavailable. Source identity also remains locally predictable in the embeddings (Supplementary Fig.~S5 and Supplementary Table~S18).

The next priority is prospective, patient-grouped evaluation across institutions, report styles and acquisition systems. Such studies should assess calibration, uncertainty, report correctness and effects on clinical workflow \cite{parasa2023framework,zhu2026gibench,jiang2026reportangel}. They should also test sensitivity to frame-selection and matching thresholds and include sufficient examples of rare findings. Similar many-frame, many-finding records occur in upper gastrointestinal endoscopy and other procedure-based imaging workflows. Extending correspondence recovery to these settings would test whether routine documentation can provide scalable vision--language supervision beyond colonoscopy.

\section{Methods}\label{method}

\subsection{Ethics and data source}
This retrospective study was approved by the Ethics Committee of Zhongshan Hospital, Fudan University (approval number B2025-145R) and was conducted in accordance with the Declaration of Helsinki. The committee granted a waiver of informed consent for the retrospective use of these routinely collected records, which were de-identified before analysis.

\subsection{Problem formulation}
Each colonoscopy case $c$ consists of a set of still frames $\mathcal{X}_c=\{x_1,\dots,x_{m_c}\}$ and a free-text report that we segment into finding sentences $\mathcal{T}_c=\{t_1,\dots,t_{n_c}\}$. Because the report is written for the case as a whole, the correspondence between a sentence and the frame that depicts it is unobserved. The initial supervision is therefore the case-level pairing $(\mathcal{X}_c,\mathcal{T}_c)$. Our aim is to learn an image encoder $f$ and a text encoder $g$ that map frames and sentences into a shared space in which a frame lies close to the sentence describing the same lesion. All embeddings are $\ell_2$-normalised, and image--text similarity is defined as $s(x,t)=\langle f(x),\,g(t)\rangle$. We progressively re-pair frames and sentences, train $f$ and $g$ on the recovered pairs with a contrastive objective, and apply the frozen or fine-tuned image encoder to downstream lesion-level tasks.

\subsection{Report corpus and curation}
The pretraining corpus comprised 280{,}476 de-identified colonoscopy reports and the still frames captured during the corresponding procedures. Of the reports, 44.2\% documented at least one polyp; the remainder described normal examinations. Each report was segmented into individual finding sentences. We parsed the anatomical site, lesion size in centimetres and a controlled vocabulary of nineteen mucosal-surface and polyp-morphology descriptors. Among the polyp-positive reports, 69.6k (56.2\%) documented a single polyp and 54.2k (43.8\%) documented two or more. A lesion size was stated in 115{,}245 polyp-positive reports (93\%); when present, the median was 0.6~cm (interquartile range, 0.4--0.8~cm). The polyp-positive cases contained 8.54 million frames (median, 65; mean, 69 per case), only some of which depicted a reported lesion. Curation removed low-quality entries and records without usable frames. The final corpus contained 104{,}542 cases and 125{,}756 image--text pairs, each linking one frame to one finding sentence.

\subsection{EndoCLIP architecture and pre-training}

\subsubsection{Dual-encoder architecture}
EndoCLIP is a CLIP-style dual-encoder model with an image encoder $f(\cdot;\theta_f)$ and a text encoder $g(\cdot;\theta_g)$. The image encoder comprises a Vision Transformer backbone \cite{dosovitskiy2021image}, $f_{\mathrm{backbone}}(\cdot)$, followed by a linear projection head $f_{\mathrm{proj}}(\cdot)$. The projection head maps the backbone output into a shared $d$-dimensional embedding space, and $\theta_f$ contains the parameters of both modules. The Transformer text encoder \cite{vaswani2017attention} maps a finding sentence into the same space through its own projection head. Both encoders are implemented in the \texttt{open\_clip} framework \cite{cherti2023reproducible}.

\subsubsection{Progressive correspondence recovery}\label{subsec:correspondence_recovery}
Pairing a report with every frame from the corresponding case yields a one-to-many or many-to-many signal in which most frame--sentence pairs are mismatched. We progressively recovered lesion-level report-to-evidence correspondence in three stages. The intermediate checkpoints are denoted EndoCLIP-I, EndoCLIP-II and EndoCLIP-III.

In the first stage (case-level evidence localisation), we sampled one frame $x_i\in\mathcal{X}_c$ at random from each case. We paired it with the generic prompt ``polyp present'' for polyp-positive cases or ``no finding'' for normal cases. Each case therefore contributed one weak pair, and no descriptive finding sentence was used. Training on this coarse signal produced EndoCLIP-I. We then used EndoCLIP-I to score every frame by its similarity to the case prompt and discarded low-scoring background or otherwise uninformative views. Training on the retained frame set $\tilde{\mathcal{X}}_c$ produced EndoCLIP-II.

In the second stage (single-lesion anchor selection), we restricted training to the 69.6k cases documenting one polyp. In each case, one finding sentence $t$ describes the lesion and therefore provides a low-ambiguity anchor. We used EndoCLIP-II to select the best-matching retained frame, $x^\star=\operatorname*{arg\,max}_{x\in\tilde{\mathcal{X}}_c}\,s(x,t)$, and formed the pair $(x^\star,t)$. Training on these lesion-level pairs produced EndoCLIP-III.

In the third stage (multi-lesion correspondence disambiguation), we applied EndoCLIP-III to the 54.2k multi-polyp cases. Because each report contributes several finding sentences, we matched sentences to frames. For every finding sentence $t_j$, we retrieved the most similar retained frame, $x^\star_j=\operatorname*{arg\,max}_{x\in\tilde{\mathcal{X}}_c}\,s(x,t_j)$. We retained $(x^\star_j,\,t_j)$ only when $s(x^\star_j,t_j)$ exceeded a fixed threshold of $0.28$, selected on a separate held-out validation set of multi-lesion cases (Supplementary Fig.~S6 and Supplementary Table~S22). Sentences without a confident frame match were discarded. This stage produced approximately 56{,}200 additional lesion-level image--text pairs, which we combined with the single-lesion pairs to train EndoCLIP.

\subsubsection{Contrastive pre-training objective}
Given a batch of $N$ paired examples $\{(x_k,t_k)\}_{k=1}^{N}$ produced by the correspondence-recovery procedure above, let $S_{kl}=s(x_k,t_l)$ denote the cosine similarity between the $k$-th frame and the $l$-th sentence. The two encoders were trained jointly with the symmetric image--text contrastive (InfoNCE) objective \cite{oord2018representation,radford2021learning}
\begin{equation}\label{eq:infonce}
\mathcal{L}=-\frac{1}{2N}\sum_{k=1}^{N}\left[\log\frac{\exp(S_{kk}/\tau)}{\sum_{l=1}^{N}\exp(S_{kl}/\tau)}+\log\frac{\exp(S_{kk}/\tau)}{\sum_{l=1}^{N}\exp(S_{lk}/\tau)}\right],
\end{equation}
where $\tau$ is a temperature parameter. The first term in equation~(\ref{eq:infonce}) aligns each frame with its matched sentence relative to the other sentences in the batch. The second aligns each sentence with its matched frame. This objective draws matched pairs together and pushes mismatched pairs apart. We used the same objective at every stage, applied to the image--text pairs available at that stage.

\subsubsection{Pre-training configuration}
At every stage, both encoders were re-initialised and trained with the \texttt{open\_clip} pipeline. Following the pipeline defaults, we used AdamW \cite{loshchilov2019decoupled} with a cosine learning-rate schedule and a 50-step linear warmup. Each contrastive stage was trained for 20 epochs in single precision (\texttt{fp32}) with a learning rate of $1\times10^{-5}$ and weight decay of $0.1$. The per-GPU batch size was 16 across two GPUs, giving a global batch size of 32. Training minimised the symmetric InfoNCE loss in equation~(\ref{eq:infonce}) with the standard learnable CLIP logit-scale temperature $\tau$, initialised at $0.07$. All contrastive stages were trained on two NVIDIA A100 80\,GB GPUs. The primary retrieval, classification and report-generation comparisons used the EndoCLIP image encoder, either frozen or fully fine-tuned as specified below. 

\subsection{Downstream evaluation}

\subsubsection{Baseline encoders}
We benchmarked EndoCLIP against general-purpose and biomedical vision--language encoders under the same downstream protocol (optimisation settings in Supplementary Table~S20). The general-purpose baselines were CLIP-OpenAI \cite{radford2021learning} and a randomly initialised CLIP model (CLIP-Scratch). The biomedical baselines were BiomedCLIP~\cite{zhang2025biomedclip}, which uses a PubMedBERT text encoder \cite{gu2021domain}, and PMC-CLIP~\cite{lin2023pmcclip}.

\subsubsection{Multi-centre evaluation cohorts}
Downstream evaluation comprised retrieval on EndoReport100, multi-centre classification and structured report generation on EndoVL. EndoReport100 was an independent in-house retrieval benchmark of 100 multi-lesion cases and 217 lesions. The classification evaluation combined five EndoVL tasks with benign-versus-malignant pathology on an independent Zhongshan Hospital cohort. EndoVL \cite{fu2026endovl} comprised nine public source datasets: CVC-ClinicDB \cite{bernal2015wm}, CVC-300 \cite{vazquez2017benchmark}, Kvasir-SEG \cite{jha2020kvasir} and six centre-specific subsets of the multi-centre PolypGen collection \cite{ali2023multi}. The images were acquired using different endoscopes and from different patient populations. From the pooled polyp images, we organised four attribute-classification tasks: four-class Paris morphology (2{,}106 images), mucosal congestion (2{,}062), lobulated surface (2{,}139) and villous surface (2{,}099). We used the expert size annotations to define a binary polyp-size task at the 10~mm threshold (2{,}177 images). The expert structured-report annotations provided the ground truth for the report-generation benchmark.

\subsubsection{Classification transfer protocols}
All six classification tasks were evaluated under zero-shot, linear-probe and end-to-end training protocols. Zero-shot prediction used task-specific clinical prompts without updating model weights (Supplementary Table~S2). Linear probing froze each image encoder and trained a single linear classifier on 5\%, 10\%, 20\%, 50\% or 100\% of the training partition. End-to-end experiments used the same label fractions and optimisation protocol for three ViT-L/14 models: EndoCLIP and a pretrained ViT-L/14 were fully fine-tuned, whereas a randomly initialised ViT-L/14 was trained from scratch (Supplementary Table~S20). 

\subsubsection{Structured report generation protocol}
The EndoVL structured-report dataset comprised 1{,}375 training, 197 validation and 394 test images. Each frozen image encoder supplied its full sequence of visual tokens to an encoder-specific trainable projector. The projector consisted of a linear layer, GELU activation, a second linear layer and layer normalisation, and mapped the visual tokens to a frozen Qwen3-14B decoder. Given a fixed instruction, the decoder generated JSON fields for lesion diameter, Paris type and a subset of nineteen descriptors. Projectors were trained for 20 epochs with target-only causal language-model loss in bfloat16, using a batch size of 4 and four-step gradient accumulation. The maximum learning rate was $1\times10^{-4}$, weight decay was 0.1, gradient clipping was 1.0. The checkpoint with the lowest validation loss was selected, and test outputs were decoded deterministically with one beam, no sampling and at most 128 new tokens.

\subsubsection{Statistical analysis}
Unless otherwise stated, label-efficient results are reported as the mean and standard deviation over three random seeds (42, 123 and 456). In the zero-shot classification heatmap, each metric is expressed as its fold change relative to the random-chance baseline for that task. The t-SNE projections \cite{maaten2008visualizing} were computed on frozen image embeddings for visualisation. We quantified embedding structure with a k-nearest-neighbour leave-one-out classifier on the same features. For EndoReport100, we quantified uncertainty with a case-level cluster bootstrap of 2{,}000 resamples. Whole cases were resampled to preserve within-case correlation and to obtain 95\% confidence intervals for each metric. Differences between EndoCLIP and each comparator were assessed with a two-sided paired bootstrap test over shared resamples, with $p<0.05$ treated as significant. For the benign--malignant reader study, we used exploratory two-sided exact McNemar tests on the shared seed-42 test images.

\section{Data availability}
The de-identified EndoReport100 benchmark, including its frames, clinician-curated frame-level polyp labels and lesion--sentence links, is publicly available on Figshare at \href{https://doi.org/10.6084/m9.figshare.33085637}{doi.org/10.6084/m9.figshare.33085637} \cite{li2026endoreport100}. De-identified restricted data may be made available by the corresponding authors for non-commercial research upon reasonable request, subject to institutional and ethical approval and an appropriate data-use agreement. The external images used to construct EndoVL originate from the public CVC-ClinicDB \cite{bernal2015wm}, CVC-300 \cite{vazquez2017benchmark}, Kvasir-SEG \cite{jha2020kvasir} and PolypGen \cite{ali2023multi} datasets and remain available under their original access terms. The derived EndoVL expert annotations and split manifests are publicly available on Figshare \cite{fu2026endovl}.

\section{Code availability}
Code to reproduce the experiments reported in this study is available at \href{https://github.com/Jia7878/EndoCLIP}{github.com/Jia7878/EndoCLIP}.

\bibliography{sn-bibliography}


\end{document}